\pdfoutput=1

\documentclass[11pt]{article}

\usepackage[preprint]{acl}

\usepackage{times}
\usepackage{latexsym}
\usepackage{amsmath} 
\usepackage{amssymb}

\usepackage[T1]{fontenc}

\usepackage[utf8]{inputenc}

\usepackage{microtype}

\usepackage{inconsolata}

\usepackage{graphicx}

\usepackage{booktabs}
\usepackage{multirow}

\usepackage{listings}
\definecolor{isarblue}{HTML}{006699}
\definecolor{isargreen}{HTML}{009966}
\lstdefinelanguage{isabelle}{%
    keywords=[1]{type_synonym,datatype,fun,abbreviation,definition,proof,lemma,theorem,corollary, qed},
    keywordstyle=[1]\bfseries\color{isarblue},
    keywords=[2]{where,assumes,shows,and,fixes},
    keywordstyle=[2]\bfseries\color{isargreen},
    keywords=[3]{if,then,else,case,of,SOME,let,in,O, define, have, use, by, using, moreover, unfolding,from,with,show},
    keywordstyle=[3]\bfseries\color{isarblue},
}
\title{Consistent Autoformalization for Constructing Mathematical Libraries}

\author{
  \textbf{Lan Zhang\textsuperscript{1}},
  \textbf{Xin Quan\textsuperscript{1}},
  \textbf{Andr\'e Freitas\textsuperscript{1,2,3}}\\
  \textsuperscript{1}Department of Computer Science, University of Manchester, United Kingdom\\
  \textsuperscript{2}Idiap Research Institute, Switzerland\\
  \textsuperscript{3}National Biomarker Centre, CRUK Manchester Institute, United Kingdom\\
  \texttt{\{lan.zhang-6, xin.quan\}@postgrad.manchester.ac.uk}\\
  \texttt{andre.freitas@idiap.ch}
}

\begin{document}
\maketitle
\begin{abstract}

Autoformalization is the task of automatically translating mathematical content written in natural language to a formal language expression. The growing language interpretation capabilities of Large Language Models (LLMs), including in formal languages, are lowering the barriers for autoformalization. However, LLMs alone are not capable of consistently and reliably delivering autoformalization, in particular as the complexity and specialization of the target domain grows. As the field evolves into the direction of systematically applying autoformalization towards large mathematical libraries, the need to improve syntactic, terminological and semantic control increases. This paper proposes the coordinated use of three mechanisms, most-similar retrieval augmented generation (MS-RAG), denoising steps, and auto-correction with syntax error feedback (Auto-SEF) to improve autoformalization quality. The empirical analysis, across different models, demonstrates that these mechanisms can deliver autoformalizaton results which are syntactically, terminologically and semantically more consistent. These mechanisms can be applied across different LLMs and have shown to deliver improve results across different model types.\footnote{Code and datasets are available at \url{https://github.com/lanzhang128/retrieval_augmented_autoformalization}}
\end{abstract}


\section{Introduction}
\begin{figure*}[!t]
    \centering
    \includegraphics[scale=0.39]{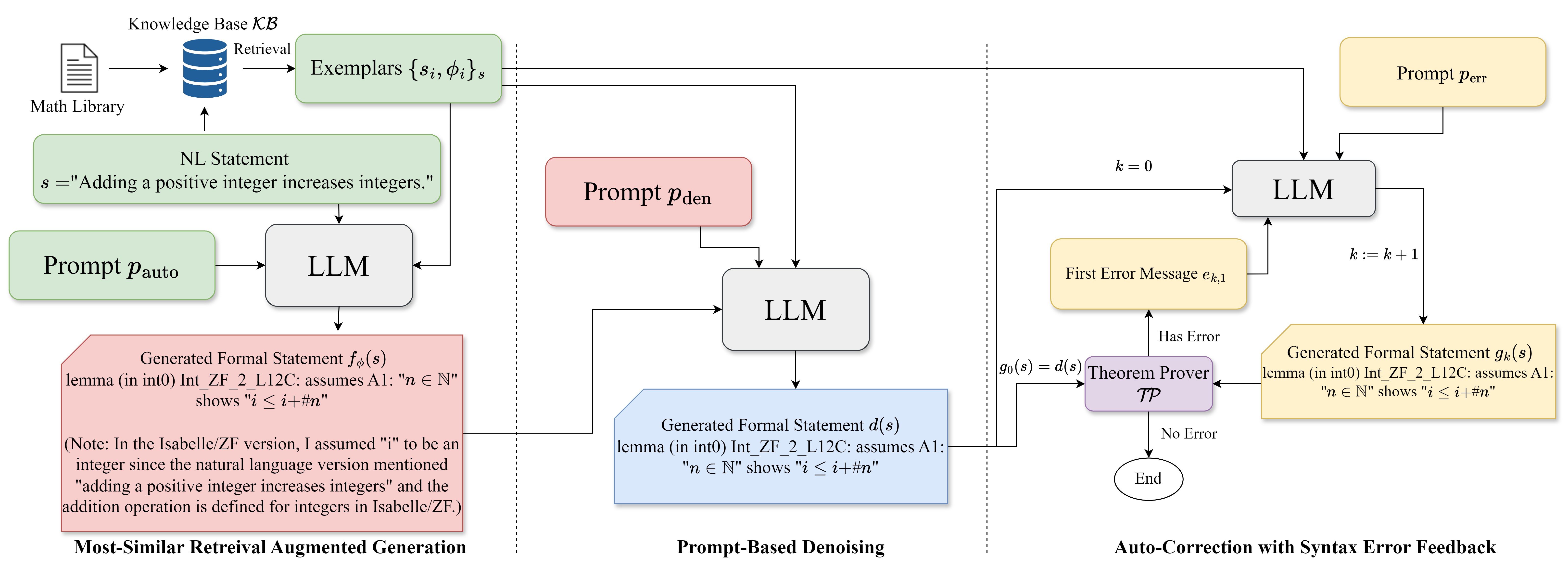}
    \caption{The overall framework consists of three stages: Stage 1 contains one round for retrieval augmented autoformalization; Stage 2 contains one round for denoising; Stage 3 is composed of several iterative rounds to refine the code based on syntax errors. For better illustration, we change \textbackslash\textless in\textgreater, \textbackslash\textless nat\textgreater, \textbackslash\textless lsq\textgreater, \$+\$ to their LaTeX version $\in$, $\mathbb{N}$, $\leq$, $+$. The ground truth code is \textit{lemma (in int0) Int\_ZF\_1\_5\_L7A: assumes "a\textbackslash\textless in\textgreater\textbackslash\textless int\textgreater"  "b \textbackslash\textless in\textgreater \textbackslash\textless int\textgreater\textbackslash\textless\^{}sub\textgreater+"shows "a \textbackslash\textless lsq\textgreater a\textbackslash\textless ra\textgreater b" "a \textbackslash\textless noteq\textgreater a\textbackslash\textless ra\textgreater b"  "a\textbackslash\textless ra\textgreater b \textbackslash\textless in\textgreater \textbackslash\textless int\textgreater"} (assumes "$a\in\mathbb{Z}$"  "$b\in\mathbb{Z}^{+}$" shows "$a\leq a+b$" "$a\neq a+b$"  "$a+b\in\mathbb{Z}$").}
    \label{fig:framework}
\end{figure*}

Mathematical reasoning constitutes an essential aspect of human intelligence~\citep{Saxton2019AnalysingMR, lu-etal-2023-survey}. It centers on symbolic-level reasoning, as manifested through systematic, abstract and and step-wise logical inference. Mathematical reasoning models has been clustered under two types: deep learning models ~\citep{hendrycks2021measuring, wei2023chainofthought, meadows-freitas-2023-introduction, Liu_Huang_Zhai_Liu_2023} and formal models~\citep{polu2020generative, wang2020learning, han2022proof, jiang2022thor,jiang2023draft}. Mathematical reasoning in Large Language Models (LLMs) predominantly uses statements expressed in informal mathematical statements. More recent models have aimed towards bridging both informal and formal mathematical reasoning ~\citep{wu2022autoformalization, First2023BaldurWG, azerbayev2023proofnet, quan-etal-2024-enhancing}, where the material (content-based) inference strengths of LLMs are complemented by external formal/symbolic reasoning methods such as automated theorem provers (e.g. Isabelle~\citep{paulson2000isabelle} and Lean~\citep{lean}),  which can systematically assess the logical validity of the reasoning process~\citep{wu2022autoformalization}, facilitating LLMs to perform controlled and consistent inference. 

However, formal and verifiable mathematical reasoning with theorem provers requires the manual formalization of logical formulae from informal statements, in order to build the supporting mathematical libraries, knowledge bases (KBs) which express previous axioms, definitions, theorems and proofs, a process that demands considerable effort and domain-specific knowledge. A prototypical case in point is the liquid tensor experiment ~\citep{doi:10.1080/10586458.2021.1926016}, an initiative aimed at formalizing analytical geometry results from Scholze \& Clausen, requiring a community coordinated effort of experts.

Contemporary LLMs have demonstrated considerable efficacy ~\citep{wu2022autoformalization, Xin2023LEGOProverNT, First2023BaldurWG} for supporting autoformalization efforts within an in-context learning paradigm, being largely evaluated in less specialized domains and tasks. Existing methods are still limited in delivering a method for systematically and consistently building large formal and specialized mathematical libraries. The essence of the challenge is twofold: (i) \textit{specialization and out-of-distribution (OOD) drifts:} as one moves towards more specialized and newer domains to be autoformalized, models are progressively exposed to more challenging OOD cases, and (ii) \textit{library consistency and coherence:} new formalized need to be consistently built-up on previously statements, cohering terminologically, syntactically and semantically.

This work targets the overarching research question: `how to systematically support the creation of consistent and coherent formal mathematical libraries from informal mathematical statements?'. In order to address this task, we decompose this broader aim into the following research questions: \textit{RQ1: `To what extent contemporary LLMs are capable of formalizing specialized mathematical statements into formal representations for mathematical libraries?'; RQ2: `Which metrics can be used to assess the quality of the formalized outputs?'; RQ3: `Which mechanisms can be used to extend the autoformalization properties of LLMs to achieve better generative control and enhance terminological, syntactic and semantic consistency and coherence?'}. To address these research questions, we propose a novel framework (See Figure \ref{fig:framework}) that leverages LLMs with most-similar retrieval augmented generation (MS-RAG), denoising steps and iterative feedback-guided syntax error refinement cycles (Auto-SEF) to deliver a syntactically consistent and semantically coherent autoformalization. 

To assess the effectiveness of our proposed framework, we construct a supporting dataset for the task of mathematical library autoformalization (MathLibForm) and build a supporting empirical analysis methodology guided by a critical selection of a set of automated metrics. We conduct a systematic empirical analysis with a diverse sample of state-of-the-art LLMs, in order to compare and contrast their autoformalization properties and the impact of the proposed library autoformalization mechanisms. Our results demonstrate that leveraging LLMs with MS-RAG and Auto-SEF, combined with denoising strategies, can significantly enhance the syntactic correctness of formalization results, reaching improvements from 5.47\% to 33.58\%. In summary, the contributions of the paper are: 
\begin{enumerate}
    \item Proposal of a novel neuro-symbolic framework targeting the autoformalization of mathematical libraries, which employs LLMs with MS-RAG, denoising and Auto-SEF to consistently and iteratively enhance and refine the formalization results; 
    \item Definition of a new task (formalization of mathematical libraries) and creation of a supporting dataset (MathLibForm);
    \item Proposal of an evaluation methodology.
\end{enumerate}

\section{Proposed Approach}\label{sec:approach}

In this section, we start by defining the target task and then describe the proposed mechanisms for improving autoformalization. 

\noindent \textbf{Autoformalization:} An autoformalization is a transformation function which maps an informal mathematical statement $s$ in the domain of natural language and LaTeX symbols $\mathcal{S}$ into a formal mathematical statement $\phi$, under a formal language $\mathcal{F}$, $f: \mathcal{S} \to \mathcal{F}$, such that for every $s \in \mathcal{S}$, there exists a $\phi \in \mathcal{F}$ where $f(s) = \phi$. 

\noindent \textbf{Semantic correctness:} A transformation $f(s) = \phi$ is semantically correct if there exists a model $\mathcal{M}$ such that: 
\[
\exists \mathcal{M} \quad \text{:} \quad \mathcal{M} \models s \quad \text{and} \quad \mathcal{M} \models \phi,
\]
where $\models$ denotes that the former item satisfies or correctly interprets the latter.

\noindent \textbf{Library-based autoformalization:} Given a Knowledge Base ($\mathcal{KB}$) of formalised mathematical statements under a formal language $\mathcal{F}$, a \textit{library-based autoformalization transformation function} $f_{\Phi}$ is defined such that the generated statement $\phi$ is semantically consistent with the set of statements $\Phi \in \mathcal{KB}$.


\noindent \textbf{Semantic consistency:} A statement \(\phi\) is semantically consistent with respect to \(\mathcal{KB}\) if all terms in \(\phi\) that have references in \(\mathcal{KB}\) are used consistently with the terms in \(\mathcal{KB}\). Formally, let \(\phi\) be a statement and \(\mathcal{KB}\) be a knowledge base. \(\phi\) is semantically consistent with respect to \(\mathcal{KB}\) if:
\[
\forall t \in \text{terms}(\phi) \cap \text{references}(\mathcal{KB}), \quad t_{\phi} = t_{\mathcal{KB}},
\]
where \(\text{terms}(\phi)\) denotes the set of terms in \(\phi\) and \(\text{references}(\mathcal{KB})\) denotes the set of referenced terms in \(\mathcal{KB}\).



\subsection{Most-Similar Retrieval Augmented Generation (MS-RAG)}

Under the aforementioned formal notations, autoformalization via LLMs defines the transformation function as:
\[
f(s)=\text{LLM}(p_\text{auto},\{(s_i,\phi_i)\},s),
\]
\noindent where $p_\text{auto}$ is a prompt for autoformalization and $\{(s_i,\phi_i)\}$ is a set of exemplars. The initial attempt~\citep{wu2022autoformalization} defined subcategories $\mathcal{SC}_j$ in math and chose fixed examples $\{(s_i,\phi_i)\}_j\in\mathcal{SC}_j$ for each subcategory, where the transformation function becomes:
\[
f(s)=\text{LLM}(p_\text{auto},\{(s_i,\phi_i)\}_j,s)\text{, if } s\in\mathcal{SC}_j.
\]

However, fixed examples cannot reflect the usage of various novel definitions and notions in each subcategory. Therefore, with the assumption of the existence of $\mathcal{KB}$, we propose to first retrieve a set of samples based on a similarity relevance function $\mathcal{MS}(s)\in\mathcal{KB}$ and then define the transformation function as:
\[
f_\phi(s)=\text{LLM}(p_\text{auto},\{(s_i,\phi_i)\}_s,s),
\]
\noindent where $(s_i,\phi_i)\in\mathcal{MS}(s)$.

\subsection{Denoising Formalization Results}
Bias inherited from instruction fine-tuning~\citep{NEURIPS2022_instruct} causes LLMs during autoformalization to occasionally generate redundant texts not integral to the formal statement, thereby infusing the final output with noisy information. Consequently, the direct output of LLMs frequently fails to meet the criteria for a valid formal code. Please note that despite the fact that output conditions can be communicated on the initial prompt, typically the output behaviour of the models cannot be fully controlled, nor fully enforceable. To alleviate this issue, we propose two types of denoising:

\noindent\textbf{Code-Based Denoising (CBD).} Definition of a set of post-processing rules $\mathcal{R}$ to remove irrelevant outputs such as \textit{extra explanations} and \textit{unsolicited proofs} , where a new formal statement is obtained: $d(s)=\mathcal{R}(f_\phi(s))$.

\noindent\textbf{Prompt-Based Denoising (PBD).} The rigidity of a CBD method can be contrasted to a post-hoc prompt-based approach for the same purpose. Hence, we propose the design of a prompt $p_\text{den}$ which performs the denoising of the autoformalization results. Denoising with only a prompt raises the risk of losing semantic consistency because of the bias in the training data of LLMs. Therefore, the set of retrieved items $\mathcal{MS}(s)$ from MS-RAG could be used to maintain semantic consistency. The denoising becomes: $d(s)=\text{LLM}(p_\text{den},\{(s_i,\phi_i)\}_s,f_\phi(s))$. 


Using reported syntax errors as a feedback have been established as a systematic mechanism for guiding the correction of formal models ~\citep{quan-etal-2024-enhancing, quan2024verification} for LLMs potentially automatically correct the formalization results. In contrast, PBD and CBD provides a template-based/prescribed mechanism for output control.


\subsection{Auto-correction with Syntax Error Feedback (Auto-SEF)}
The validity of a formal code $\phi$ can be checked by a theorem prover $\mathcal{TP}$ that supports its written formal language $\mathcal{F}$. If the formal code is not valid, the theorem prover can output a set of syntax errors $\{e_k\}=\mathcal{TP}(\phi)$. Using reported syntax errors as feedback has been established as a systematic mechanism for guiding the correction of formal models ~\citep{quan-etal-2024-enhancing, quan2024verification}, potentially allowing LLMs to automatically correcting the results of formalization. Hence, we design a prompt $p_\text{err}$ to add an auto-correction component to let LLMs recognize previously produced errors and correct mistakes. To maintain semantic consistency, retrieved examples are also used and the generation becomes:
\[
g(s)=\text{LLM}(p_\text{err},\{(s_i,\phi_i)\}_s,\{e_k\},d(s)).
\] 
\noindent where $\{e_k\}=\mathcal{TP}(d(s))$. Within this setting we propose an iterative process:

\[g_{k+1}(s)=\text{LLM}(p_\text{err},\{(s_i,\phi_i)\}_s,e_{k,1},g_k(s))\]
with initial state $g_0(s)=d(s)$ and $e_{k,1}$ is the first item in $\mathcal{TP}(g_k(s))$.


\section{Evaluation Benchmark}\label{sec:data}

\subsection{MathLibForm}

Formal mathematical datasets, such as miniF2F \citep{zheng2022miniff}, predominantly concentrate on distinct mathematical problems representing simpler mathematical solving tasks. In contrast, the creation of mathematical libraries demands the autoformalization of statements which can be more specialized, conceptually more complex and potentially out-of-distribution. In this work we use  \textit{IsarMathLib}\footnote{\url{https://github.com/SKolodynski/IsarMathLib}}, as a reference setting within the environment of the Isabelle/ZF theorem prover framework. Formal statements in IsarMathLib are frequently accompanied by textual comments, which serve as the corresponding natural language statements of the formal expressions. Mathematical items, such as \textit{lemma}, \textit{definition}, \textit{corollary}, \textit{theorem}, along with textual comments and proofs, were first systematically extracted via a script. This led to a total of 2,744 items, which were then randomly divided into training and test sets in a 90\% to 10\% split, resulting in 2,470 training samples and 274 test samples for constructing the MathLibForm dataset. To enrich the information contained in MathLibForm, we also informalize formal statements with Mistral and add the generated textual descriptions. The training and testing sets are used to define the knowledge base $\mathcal{KB}$ and the evaluation.

\subsection{Evaluation Metrics}
Assessing the overall correctness of autoformalized code outputs requires resource intensive specialized/expert-level human feedback. In addition, human evaluations can become largely subjective in situations where the assessment criteria is too complex to be elicited (i.e., the validation process cannot be systematized into a protocol), which, we argue, is the case for autoformalization. The dependency on multiple human validators with skills in both theorem provers and the underlying multi-domain mathematics makes this problem particularly severe.

We mentioned semantic correctness and consistency in Section~\ref{sec:approach} as the final desirable properties as an outcome of an autoformalization process, which can become too strict for evaluating autoformalization tasks with current LLMs. Therefore, in this work, we propose two distinct proxies to assess code correctness: \textit{semantic similarity} and \textit{syntactic correctness}. Utilizing the ground truth as a reference, we measure semantic similarity using pairwise metrics, including BLEU~\citep{papineni-etal-2002-bleu}, ChrF~\citep{popovic-2015-chrf}, RUBY~\citep{RUBY}, and CodeBERTScore (CBS)~\citep{zhou-etal-2023-codebertscore}. The description of these metrics are provided in the \textit{Appendix}. For syntactic correctness, we use the Isabelle theorem prover to detect syntax errors in formal statements and use the \textit{Pass} metric which represents the success rate at which the generated formal statement does not exhibit any syntax errors, as verified by the theorem prover. The integration between the transformer and Isabelle is done on a ToolFormer setting with the support of an Isabelle client\footnote{\url{https://github.com/inpefess/isabelle-client}}~\citep{shminke2022python}.

\begin{table*}[!t]
    \centering
    \begin{tabular}{l p{0.27\textwidth} c c c c c}
        \toprule
        LLM & Method & BLEU-2 & ChrF & RUBY & CBS & Pass\\
        \hline
        \multicolumn{7}{l}{\textit{Baselines}}\\
        \hline
        Mistral & Zero-Shot & 0.30 & 17.14 & 16.13 & 51.13 & 0.0\\
        \hline
        Mistral & 3-Shot & 1.77 & 27.30 & 24.02 & 62.73 & 5.47\\
        \hline
        Llemma 7B & Zero-Shot & 0.91 & 16.67 & 14.77 & 47.74 & 9.12\\
        \hline
        Llemma 7B & 3-Shot & 2.43 & 28.81 & 21.93 & 66.68 & 8.76\\
        \hline
        Mixtral & Zero-Shot & 0.65 & 16.33 & 17.97 & 51.07 & 0.36\\
        \hline
        Mixtral & 3-Shot & 5.37 & 30.53 & 28.51 & 62.86 & 1.09\\
        \hline
        GPT-3.5-Turbo & Zero-Shot & 2.15 & 17.81 & 21.93 & 51.69& 40.51\\
        \hline
        GPT-3.5-Turbo & 3-Shot & 14.23 & 37.95 & 39.13 & 67.26 &38.69 \\
        \hline
        \multicolumn{7}{l}{\textit{Retrieval Augmented Autoformalization}}\\
        \hline
        Mistral & Query: \textbf{T} Index: \textbf{T} & 10.05 & 51.38 & 44.82 & 76.93 & 21.53\\
        \hline
        Mistral & Query: \textbf{T} Index: \textbf{T}+\textbf{S} & 9.96 & 50.79 & 43.92 & 76.21 &  19.71\\
        \hline
        Mistral & Query: \textbf{T} Index: \textbf{I}+\textbf{S} & 5.65 & 36.92 & 32.23 & 67.47 & 8.76\\
        \hline
        Mistral & Query: \textbf{T} Index: \textbf{T}+\textbf{I}+\textbf{S} & 10.53 & 49.61 & 43.28 & 75.17 & 22.26\\
        \hline
        Mistral & Query: \textbf{T}+\textbf{ZS} Index: \textbf{T} & 10.14 & 46.89 & 40.76 & 73.69 & 12.77\\
        \hline
        Mistral & Query: \textbf{T}+\textbf{ZS} Index: \textbf{T}+\textbf{S} & 8.40 & 46.26 & 39.91 & 73.40 & 14.96\\
        \hline
        Mistral & Query: \textbf{T}+\textbf{ZS} Index: \textbf{I}+\textbf{S} & 5.51 & 36.71 & 31.94 & 66.91 & 10.95\\
        \hline
        Mistral & Query: \textbf{T}+\textbf{ZS} Index: \textbf{T}+\textbf{I}+\textbf{S} & 8.85 & 45.14 & 39.27 & 72.47 & 16.06\\
        \hline
        Llemma 7B & Query: \textbf{T} Index: \textbf{T} & 4.18 & 36.93 & 28.68 & 69.93 & 12.77\\
        \hline
        Llemma 7B & Query: \textbf{T} Index: \textbf{T}+\textbf{S} & 4.61 & 37.48 & 29.39 & 69.56 & 14.23\\
        \hline
        GPT-3.5-Turbo & Query: \textbf{T} Index: \textbf{T} & 36.32 & \textbf{59.63} & \textbf{58.51} & \textbf{79.14} & \textbf{64.60}\\
        \hline
        GPT-3.5-Turbo & Query: \textbf{T} Index:  \textbf{T}+\textbf{S} & \textbf{37.11} & 58.56 & 57.71 & 78.89 & 62.77\\
        \bottomrule
    \end{tabular}
    \caption{Autoformalization results for different settings. BM25 retriever is used to retrieve Top-3 most similar samples for retrieval augmented autoformalization. Greedy decoding is used in generation for reproducibility. Code-based denoising is applied to all outputs. The query used to retrieve relevant exemplars includes: (\textbf{T}): natural language textual description; (\textbf{ZS}): zero-shot autoformalization result from Mistral. The index used for knowledge base has the following options: (\textbf{T}): natural language textual description; (\textbf{I}): informalization of formal statement generated from Mistral; (\textbf{S}): formal statement. The setting with highest scores is highlighted in \textbf{bold}.}
    \label{tab:bleu}
\end{table*}
\section{Experiments and Analysis}\label{sec:exp}

\subsection{Retrieval Augmented Autoformalization}
We establish baselines in zero-shot and 3-shot settings on several state-of-the-art LLMs: Mistral~\citep{jiang2023mistral}, Llemma 7B~\citep{azerbayev2024llemma}, Mixtral~\citep{jiang2024mixtral}, GPT-3.5-Turbo (descriptions of the models can be found in the \textit{Appendix}). The inclusion criteria for the selected foundation models prioritized: (i) sample diversity across the three modalities (model size, type, and specialization level), leading to 4 baseline foundation models; and (ii) a priority on open models, where the underlying modeling strategies are more transparent.

For MS-RAG, BM25~\citep{BM25} is used as the primary ranking function to retrieve Top-k (k=3) most similar samples for exemplars (BM25 will concentrate a terminological similarity function). Different settings are contrasted for querying and indexing the reference KB. There are two choices for query: 1. natural language textual description; 2. description along with zero-shot autoformalization result from Mistral. The choices for indexing KB elements combine three content sources: 1. natural language textual description; 2. informalization of formal statements; 3. formal statements. For this specific analysis, we constrain the foundation model to Mistral. All results are reported in Table~\ref{tab:bleu}.



\noindent\textbf{MS-RAG can improve autoformalization in mathematical libraries settings.} 
As shown in Table~\ref{tab:bleu}, for the same type of LLMs, using retrieved examples rather than fixed examples leads to an improvement in both semantic similarity and syntactic correctness of the generated formal statements. This mechanism can lift the performance of smaller models: e.g. as a smaller model, Mistral (7B) with MS-RAG can outperform Mixtral (8$\times$7B) with standard prompting across all metrics and is comparable to GPT-3.5 (175B) without MS-RAG according to some metrics such as RUBY.

\begin{table*}[!t]
    \centering
    \begin{tabular}{l c c c c c}
        \toprule
        Metric & MS-RAG & PBD 1A & PBD 1B & PBD 1C & PBD 1D\\
        \hline
        BLEU-2 & 6.33 (+3.72) & 8.88 (+1.61) & 11.30 (+1.99) & 15.21 (+1.49) & 14.90 (\textbf{+2.42})\\
        ChrF & 48.45 (\textbf{+2.93}) & 38.27 (-0.35) & 43.25 (-0.06) & 44.52 (-0.23) & 48.51 (+0.11)\\
        RUBY & 28.99 (+15.83) & 38.23 (+2.12) & 42.08 (+1.91) & 44.59 (+0.79) & 46.43 (\textbf{+0.98})\\
        CBS & 76.40 (\textbf{+0.53}) & 68.04 (-0.03) & 70.51 (-0.07) & 71.92 (+0.01) & 74.07 (+0.03)\\
        Pass & 17.15 (+4.38) & 6.57 (+0.00) & 9.12 (+0.00) & 13.50 (+0.37) & 28.10 (\textbf{+0.00})\\
        \bottomrule
    \end{tabular}
    \caption{The effect of denoising on Mistral. The change of scores after applying CBD is recorded in round brackets. The setting with highest final scores is marked in \textbf{bold}.}
    \label{tab:round1}
\end{table*}

\noindent\textbf{Similarity-based few-shot outperforms zero-shot learning.} For all LLMs, autoformalization results with 3-shot exemplars are generally better than those from the zero-shot setting in terms of semantic similarity metrics. For syntactic correctness, Llemma 7B and GPT-3.5 in the zero-shot setting have slightly higher pass rates compared to the 3-shot setting. 

\noindent\textbf{MS-RAG levels the playing field across models of different scales.} As the largest LLM in this experimental setting, GPT-3.5 with MS-RAG significantly outperforms all other models. However, comparing its best performance with MS-RAG to its performance in the 3-shot setting, its relative change in syntactic correctness (67\%) is much lower than that with Mistral (307\%). The relative change for Llemma 7B is the smallest (62\%). We attribute this to the fact that Llemma was not fine-tuned with instructions. These differences suggest that smaller LLM with instruction tuning benefits more from RAG.


\begin{table*}[!t]
    \centering
    \begin{tabular}{p{0.15\textwidth} p{0.75\textwidth}}
        \toprule
        NL Statement &  The next lemma shows that open sets are members of (what we will prove later to be) the natural neighborhood system on $X=\bigcup T$.\\
        \hline
        Ground Truth &
        \begin{lstlisting}[aboveskip=-10pt,belowskip=-12pt]
lemma open_are_neighs: assumes "U\<in>T" "x\<in>U"
  shows "x \<in> \<Union>T" and "U \<in> {V\<in>Pow(\<Union>T)
  .\<exists>U\<in>T. (x\<in>U \<and> U\<subseteq>V)}"
        \end{lstlisting}\\
        \hline
        3-shot & \begin{lstlisting}[aboveskip=-10pt,belowskip=-12pt]
lemma open_set_in_nhs: assumes "X = \<bigcup> T" "T \<in>
  Covers(X)" "U \<in> Open(X)"shows "U \<in> NHS(X)"
where NHS(X) is a predicate representing the natural 
neighborhood system on X. Note: The definition of NHS(X) 
should be provided before using this lemma.
        \end{lstlisting}\\
        \hline
        MS-RAG & 
        \begin{lstlisting}[aboveskip=-10pt,belowskip=-12pt]
lemma open_is_neighbors: assumes "U \<in> T" and "x \<in> U" 
  shows "U \<in> ({neighborhood system of} T)`(x)"
(* Note: This lemma assumes that T is a topology, but it is
not explicitly stated in the natural language version.*)
        \end{lstlisting}\\
        \hline
        PBD 1A & 
        \begin{lstlisting}[aboveskip=-10pt,belowskip=-12pt]
lemma open_is_neighbors: assumes "U :: set T" and "x :: T" 
  shows "U :: ({neighborhood system of} T) x"
        \end{lstlisting}\\
        \hline
        PBD 1B & 
        \begin{lstlisting}[aboveskip=-10pt,belowskip=-12pt]
lemma open_is_neighbors: assumes "U \<in> T" and "x \<in> U" 
  shows "U \<in> ({neighborhood\_system} T)`(x)"
        \end{lstlisting}\\
        \hline
        PBD 1C & 
        \begin{lstlisting}[aboveskip=-10pt,belowskip=-12pt]
lemma open_is_neighbors: assumes "U \<in> T" "x \<in> U" 
  shows "U \<in> ({neighborhood system of} T) x"
        \end{lstlisting}\\
        \hline
        PBD 1D & 
        \begin{lstlisting}[aboveskip=-10pt,belowskip=-12pt]
lemma open_is_neighbors: assumes "U \<in> T" and "x \<in> U" 
  shows "U \<in> ({neighborhood system of} T)`(x)"
        \end{lstlisting}\\
        \bottomrule
    \end{tabular}
    \caption{An example using Mistral shows that only MS-RAG and PBD 1D have no syntax errors of formalization.}
    \label{tab:case1}
\end{table*}

\noindent\textbf{Augmenting the index with auto-informalization or the query with zero-shot auto-formalization does not lead to better retrieval.} Among all results in Table~\ref{tab:bleu}, GPT-3.5 with textual description query and textual description index achieves highest scores in four metrics except BLEU-2. This query and index combination setting is also an optimal choice for Mistral, as this setting leads to highest scores in ChrF, RUBY and CBS and second highest scores in Pass. Incorporating zero-shot results from Mistral as queries generally yields worse results compared to its counterpart. This is probably caused by the low quality of zero-shot formalization results. The application of informalized descriptions during indexing also does not lead to a performance improvement.






\subsection{Output Denoising}
In this section, we investigate the impact of denoising. We select the result of MS-RAG (Query: T, Index: T) to apply PBD with four prompts: (\textbf{1A}) The prompt only contains instructions to remove explanations and proofs; (\textbf{1B}) 1A adds an additional instruction for \textit{stylistic alignment} to declare that the final output after refinement should maintain the same syntactic style; (\textbf{1C}) Includes some fixed formal statement examples for the stylistic alignment instruction in 1B; (\textbf{1D}) Changes the fixed examples in 1C to retrieved examples from MS-RAG. We record the results of Mistral in Table~\ref{tab:round1}.

\noindent\textbf{Denoising significantly impacts the quality of the formal statements.} Compared to results without denoising, using either denoising method can significantly improve BLEU and RUBY scores. Applying CBD to the original MS-RAG results can lead to an improvement across metrics. However, the effect of CBD decreases after we apply PBD to the results. For the Pass metric, performing CBD after PBD had no observable impact. This demonstrates the impact of PBD as a syntactic control mechanism. Our results suggest that a composition of PBD and CBD can yield the best performance in syntactic correctness while maintaining semantic similarity at the same or higher level.

\noindent\textbf{Denoising can reduce the performance gap between smaller and larger LLMs.} We also conducted similar experiments on GPT-3.5 (results in \textit{Appendix}). Denoising methods have a comparatively lower effect on the results of GPT-3.5, serving more as a function of control for smaller models, approaching their performance to larger models. 

\noindent\textbf{Stylistic alignment is necessary when applying PBD.} Without the explicit declaration of stylistic alignment (1A), the syntactic correctness drops 10.58\% compared to the results of MS-RAG. The reason is that when only prompted to remove redundant strings, some models tend to neglect the original syntactic style of the formal statements and rewrite them in the style that it was trained on. However, merely specifying that the model should maintain such a style without explicit examples (1B) does not effectively communicate the intent to preserve the style. This is demonstrated by the higher performance of 1C compared to 1B. In addition, using retrieved examples (1D) rather than fixed examples (1C) can further improve the results. 

\paragraph{Case Study} The example in Table~\ref{tab:case1} communicates the necessity of denoising. As shown in Table~\ref{tab:case1}, both 3-shot and MS-RAG results include an additional textual description in the final output which does not form a formal statement.  PBD 1A changes ``\textbackslash<in>'' into ``::'' which is another way of expressing ``$\in$'' but this expression is not provided in its prompt, so this behaviour is highly likely to be an inherited bias. PBD 1B and 1C mitigate this behaviour but they also introduce other syntactic errors, such as the missing word ``of'' or the special character ``\`{}''. \textbf{Only PBD 1D maintains the validity of the formal statement because the similarity of the retrieved examples and are thus emphasized during generation.}


\subsection{Iterative Symbolic Refinement}

\begin{figure}[!t]
    \centering
    \includegraphics[scale=0.15]{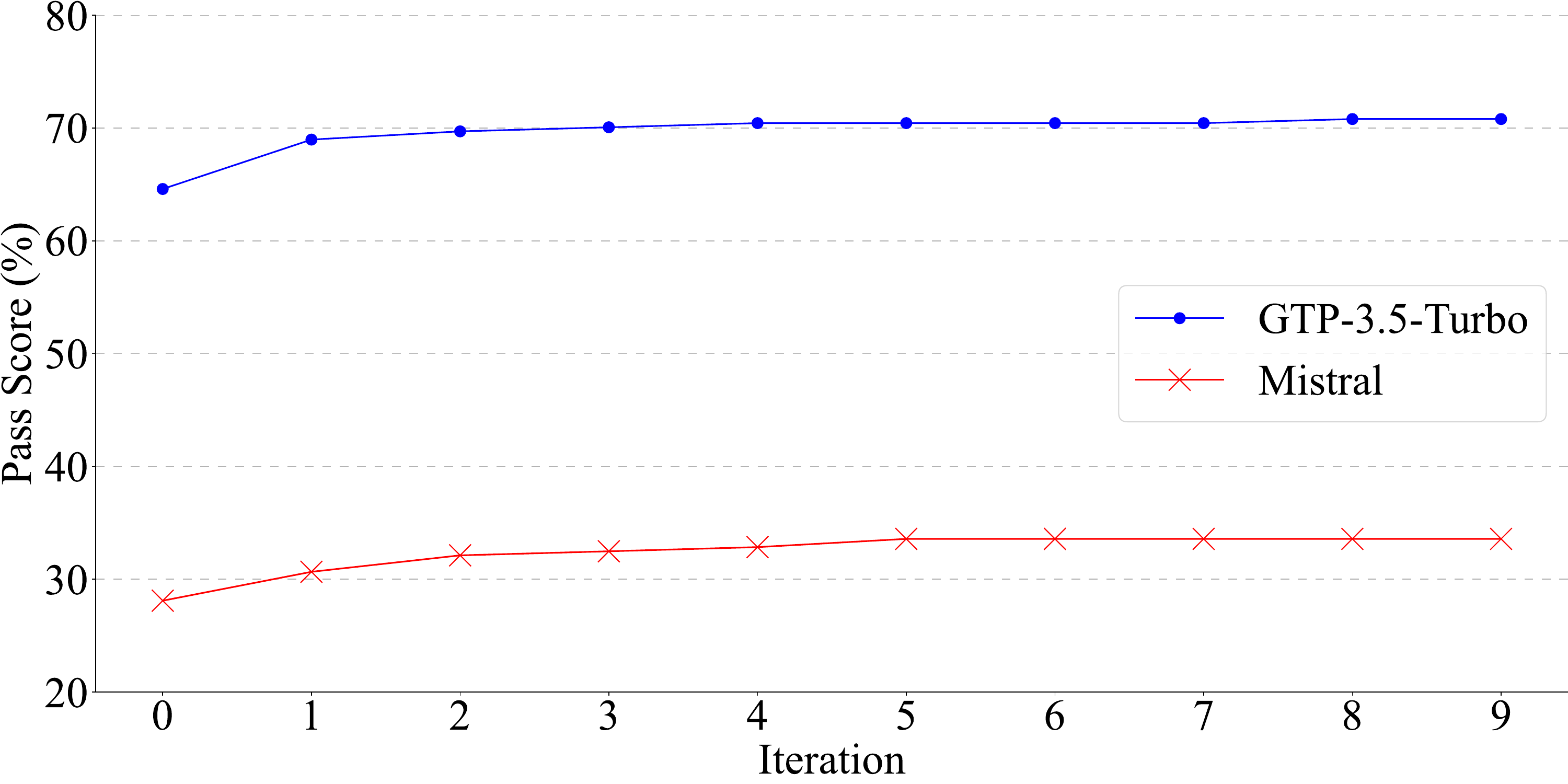}
    \caption{Pass rate of each iteration with Auto-SEF. Iteration 0 is the start point before applying Auto-SEF.}
    \label{fig:pass}
\end{figure}

In this section, we mainly focus on answering the question on whether syntactic errors can be corrected by LLMs in coordination with symbolic solvers. This process is iteratively run for up to nine cycles. To better illustrate the changes, we plot the scores of each iteration on the Pass metric in Figure~\ref{fig:pass}.

\begin{figure}[!t]
    \centering
    \includegraphics[scale=0.15]{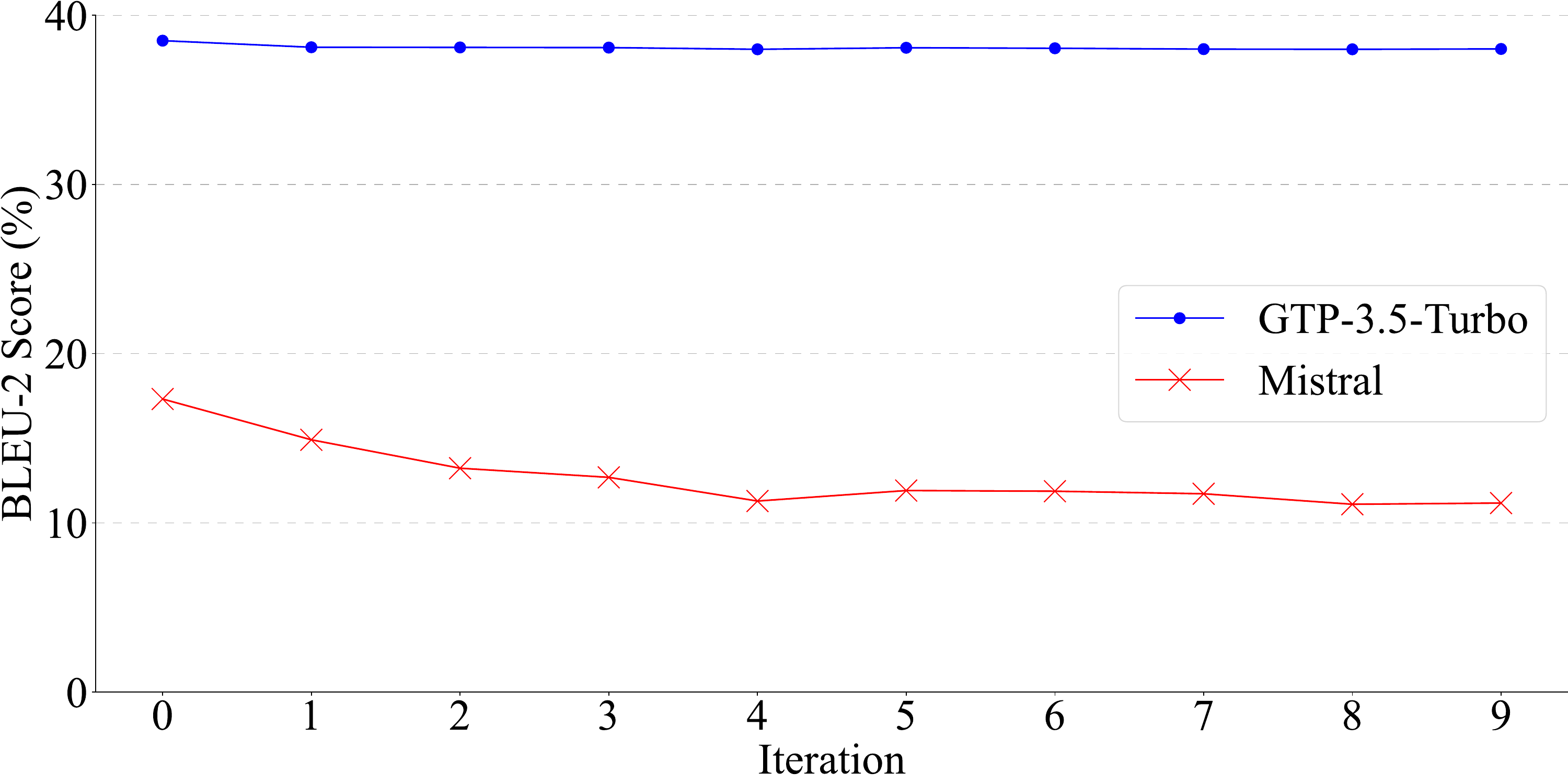}
    \caption{BLEU-2 scores of each Auto-SEF iteration.}
    \label{fig:bleu}
\end{figure}

\noindent\textbf{Iterative Auto-SEF improves syntactic correctness of the formalization results.} As shown in Figure~\ref{fig:pass}, both GPT-3.5 and Mistral can receive improvements from the iterative Auto-SEF method. This result demonstrates that Auto-SEF can indeed enable LLMs to fix some syntactic errors. The first iteration brings the largest increase (2.56\% for Mistral, 4.38\% for GPT-3.5) in pass rate. After that, the change becomes smoother and iterative improvements are limited to a small number of cycles. 

\noindent\textbf{Smaller LLM tends to trade-off semantic similarity for syntactic correctness when applying Auto-SEF.} We focus on BLEU-2 as a proxy for semantic similarity and illustrate the scores of each iteration in Figure~\ref{fig:bleu}. The BLEU-2 scores for GPT-3.5 remain steady across different iterations, whereas for Mistral, the scores decrease in the first few iterations. Combining this result with the improvement in pass rate, we hypothesize that a trade-off occurs due to the comparatively lower capacity of the model to perform syntactic correction while controlling for semantic drifting during the Auto-SEF prompting.



\subsection{A Critique of the Metrics}

\begin{figure}[!t]
    \centering
    \includegraphics[scale=0.28]{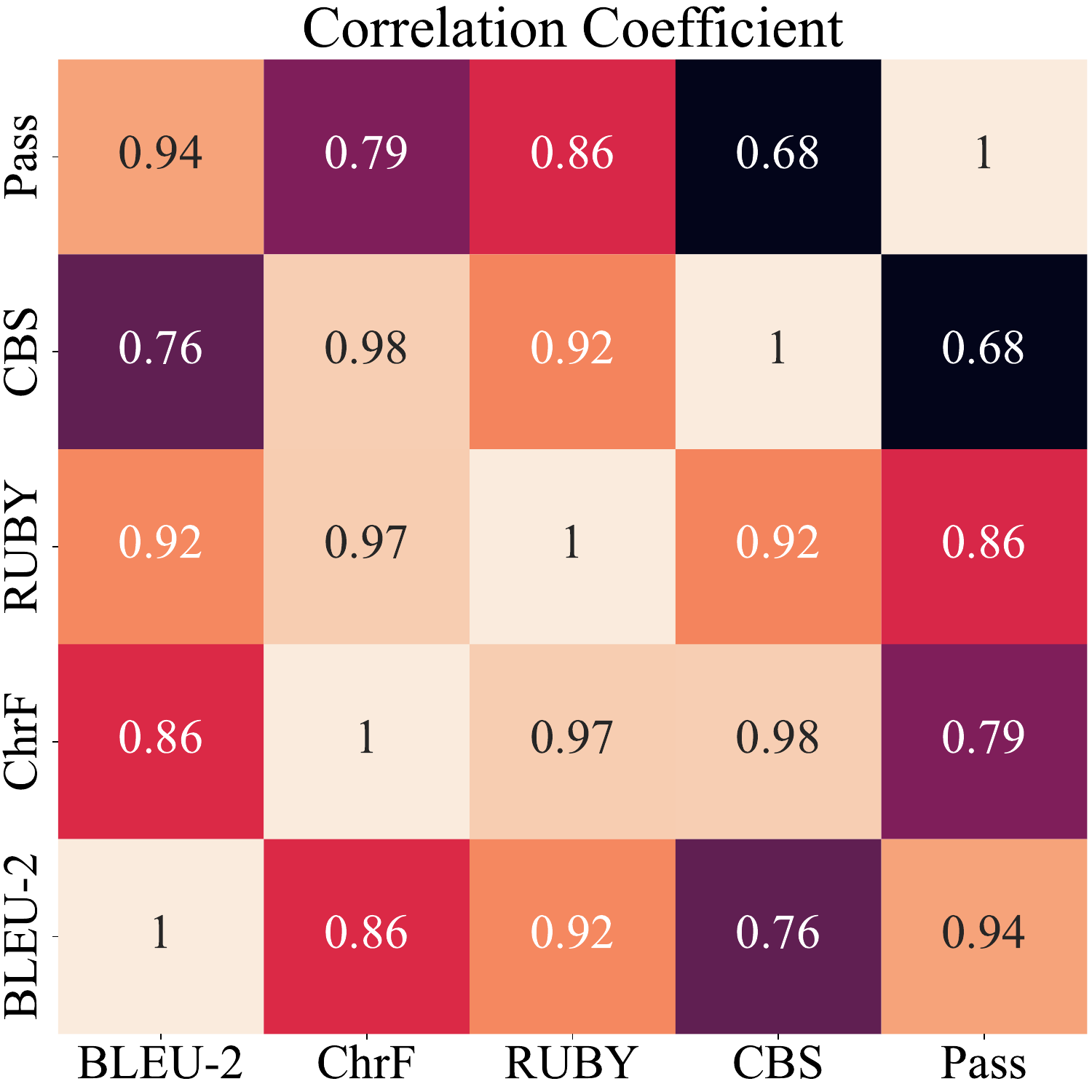}
    \caption{Correlation coefficients between metrics.}
    \label{fig:corr}
\end{figure}

Evaluation metrics for autoformalization can disagree with each other~\citep{code-generation-metrics}. We use all the results under CBD to calculate the Pearson product-moment correlation coefficients across the metrics, illustrating these coefficients with a heatmap in Figure~\ref{fig:corr}.

\noindent\textbf{RUBY can serve as an initial metric when evaluating formalization results.} All correlation coefficients are larger than 0.6. This suggests that all metrics are positively related to each other and that any one of them is a reasonable indicator for evaluating formalization results. Among these metrics, RUBY has the strongest correlation ($>0.85$) with the other metrics. 

\noindent\textbf{Pass and BLEU metrics should be jointly used to prevent evaluation bias.} Some zero-shot results in Table~\ref{tab:bleu} lead to a high score on the Pass metric but lower scores on other metrics due to internal LLM style biases. According to Figure~\ref{fig:corr}, among metrics for semantic similarity, BLEU-2 has the strongest correlation with the Pass metric and hence can indicate syntactic correctness to some extent. We suggest considering both BLEU scores and Pass rate when comparing results.

\section{Related Work}


\paragraph{Autoformalization} Autoformalization bridges the gap between natural language and formal language. \citet{cunningham-etal-2022-towards} trained a standard transformer for proof autoformalization in Coq. \citet{lu2024process} proposed a process-driven framework for autoformalization in Lean4. With the increased inference capabilities of LLMs in recent years, \citet{wu2022autoformalization,jiang2023draft} employ LLMs to autoformalize mathematical contents to Isabelle/HOL. Building on this foundation, our work also focuses on the improvement of autoformalization capabilities over LLMs.

\paragraph{Retrieval Augmented Generation~\citep{lewis2020rag}}
RAG has demonstrated improvements for code generation~\citep{lu-etal-2022-reacc, zhang-etal-2023-refsql} and in formal settings, \citet{yang2023leandojo} trained a retrieval-augmented language model for formal premise selection and theorem proving. Meanwhile, our work focuses on utilizing RAG for the task of improving autoformalization performance and coherence with respect to mathematical libraries. 




\paragraph{LLMs Refinement} Through feedback-guided refinement strategies LLMs can perform self-correction~\citep{pan-etal-2024-automatically}. Recent studies~\citep{madaan2023selfrefine, quan-etal-2024-enhancing} evaluate strategies using iterative feedback to refine LLM-generated answers for downstream tasks. Previous work has used error messages generated by theorem provers as a mechanism to support  the interface between LLMs and formal models~\citep{pan-etal-2023-logic, quan-etal-2024-enhancing, jiang2024leanreasoner, quan2024verification} and also repair models~\citep{First2023BaldurWG} to address syntactic or proof errors. Similarly, our work applies prompt-based refinement from external feedback error messages generated by Isabelle/ZF to iteratively refine the formalized logical forms with specific error code locations.


\section{Conclusion}
This paper examined the effects of using RAG for autoformalization with LLMs and explored methods to refine formalization results. Our experiments demonstrated the effectiveness of incorporating a retrieval process for autoformalization. Further experiments indicated that denoising and iteratively refining syntax errors can enhance autoformalization quality. We evaluated results on different LLMs and found that smaller LLMs with instruction fine-tuning benefited more from the proposed methods, pointing in the direction of serving as a mechanism for reducing the formal performance gaps between larger and smaller models. We also built a dataset and assessed metrics for evaluating autoformalization, which could serve as resources/methodological contributions for formal mathematical reasoning tasks. We believe combining the semantic similarity metrics with the syntactic correctness metric is a reasonable proxy for semantic correctness. 



\section*{Acknowledgements}
This work was partially funded by the Swiss National Science Foundation (SNSF) project NeuMath (\href{https://data.snf.ch/grants/grant/204617}{200021\_204617}).

\section*{Limitations}
Some natural language statements in our dataset are too general or informal, failing to provide a substantial content for autoformalization. Although our proposed framework, Auto-SEF, enhances syntactic control in autoformalization, increasing iterations do not yield significant improvements in the Pass metric. This limitation is due to the inability of LLMs to generate syntactically correct complex formal representations. 


\bibliography{ref}

\appendix

\section{Large Language Models}
We describe the large language models used in our experiments in this section.

\noindent\textbf{Mistral~\citep{jiang2023mistral}} Mistral is a large language model with 7 billion parameters which is fine-tuned on instruction datasets. It leverages several techniques such as Sliding Window Attention to boost model efficiency. To the best of our knowledge, it is also the strongest model in the domain of code and mathematics at this size. 

\noindent\textbf{Llemma~\citep{azerbayev2024llemma}} Llemma is an open large language model trained specifically for mathematics. It is pre-trained on Proof-Pile-2 which is a diverse mixture of math-related textual and code content. However, it has not been trained to follow instructions. Llemma has two scales in 7B and 34B. We only use the 7B model in our experiments.

\noindent\textbf{Mixtral~\citep{jiang2024mixtral}} Mixtral is a large language model with sparse mixture of experts method and instruction fine-tuning. It has the same architecture as Mistral 7B but each layer of it consists of 8 feed-forward blocks, making it a 8$\times$7B size model. However, during inference, only 13B parameters are activated. 

\noindent\textbf{GPT-3.5-Turbo} GPT-3.5-Turbo is a large language models of OpenAI GPT-3.5 series. It shares the same architecture as GPT-3~\citep{NEURIPS2020_gpt3} and is instruction fine-tuned. The number of parameters in GPT-3.5-Turbo is 175 billion.

\section{Evaluation Metrics}
We describe the implementation of metrics to measure semantic similarity in this section.

\noindent\textbf{BLEU~\citep{papineni-etal-2002-bleu}} The autoformalization task is a type of translation task so the most common metric in translation tasks, BLEU, is used as one evaluation metric for autoformalization. This metric is also used in \citep{wu2022autoformalization} within the same context. An implementation from NLTK~\citep{bird-loper-2004-nltk} is used.

\noindent\textbf{ChrF~\citep{popovic-2015-chrf}} ChrF is another n-gram metric in translation task that focuses on characters instead of words in BLEU. We leverage this character level metric in NLTK to take character-level granularity into account.

\noindent\textbf{RUBY~\citep{RUBY}} The autoformalization task is also a code generation task. RUBY is a metric designed specifically for code generation evaluation and uses edit distance to calculate the similarity score. If program dependence graph (PDG) or abstract syntax tree (AST) is provided, it calculates graph similarity based on graph edit distance or tree similarity based on tree edit distance. Otherwise, it calculates string edit distance to determine the string similarity between the reference code and candidate code as the score. In our experiments, because of the difficulty of obtaining PDG or AST of formal statements, we use string edit distance from NLTK to calculate string similarity as the score. This implementation focuses on characters rather than tokens as in the original paper but it still makes the score a reasonable indicator of performance.

\noindent\textbf{CodeBERTScore~\citep{zhou-etal-2023-codebertscore}} CodeBERTScore is a model-based metric to evaluate performance on code generation. It uses token representations of reference code and candidate code to determine a final score. The original paper trained different models for different programming languages to get representations: however Isabelle is not within this scope. Therefore, we use a mathematical specific model, Llemma 7B~\citep{azerbayev2024llemma}, as the supporting representation model. Although this model is not a BERT-based model, it can still generate meaningful representations for score calculation.

\section{Prompts}\label{app:prompt}
We provide prompts for informalization, autoformalization, denoising, and Auto-SEF in Table~\ref{tab:prompt_inf}, \ref{tab:prompt_auto}, \ref{tab:prompt_den}, \ref{tab:prompt_sef}, respectively.

\begin{table}[t]
    \centering
    \begin{tabular}{p{0.45\textwidth}}
        \toprule
        Translate the following Isabelle/ZF code: \newline\{statement\}\newline into a natural language version statement as brief as possible: \\
        \bottomrule
    \end{tabular}
    \caption{Prompt for informalization.}
    \label{tab:prompt_inf}
\end{table}

\begin{table}[t]
    \centering
    \begin{tabular}{p{0.45\textwidth}}
        \toprule
        Natural language version: \{Natural Language Text\} \newline Translate the natural language version to an Isabelle/ZF version without any additional text and do not give any proof: \{Formal Statement\}\\
        \bottomrule
    \end{tabular}
    \caption{Prompt for autoformalization.}
    \label{tab:prompt_auto}
\end{table}

\begin{table*}[t]
    \centering
    \scalebox{0.85}{
    \begin{tabular}{p{0.1\textwidth} p{\textwidth}}
        \toprule
         & Prompt\\
        \hline
        PBD 1A & You are an expert in Isabelle theorem prover. You will be provided with an Isabelle/ZF code generated by a language model. Your task is to clean the provided Isabelle/ZF code with following instructions. Instructions:
        \newline 1. The provided code might contain several lemmas or definitions or theorems. The cleaned code must only keep the best one lemma or definition or theorem.
        \newline 2. Do not write any proof and if there is a proof in the provided code, remove it from the cleaned code.
        \newline 3. You should only output tokens that compose the cleaned code. Anything else, including but not limited to note, description, explanation and comment, must be removed from the final answer. Giving any additional text is prohibited.
        \newline Strictly follow the instructions that I have claimed.
        \newline Provided Isabelle/ZF Code: \{isabelle code\}
        \newline Cleaned Code: \\
        \hline
        PBD 1B & 1A + An additional instruction:
        \newline 4. The cleaned code must have the same style and usage of operators as the original provided code. Operators usually start with ``\textbackslash'' such as ``\textbackslash\textless in\textgreater'', ``\textbackslash\textless cdot\textgreater''.\\
        \hline
        PBD 1C & 1A + An additional instruction:
        \newline 4. The cleaned code must have the same style and usage of operators as the original provided code. Operators usually start with ``\textbackslash'' such as ``\textbackslash\textless in\textgreater'', ``\textbackslash\textless cdot\textgreater''. Here are some additional Isabelle/ZF code examples which have the same style as the original provided code:
        \newline\{fixed 3-shot formal statements\}\\
        \hline
        PBD 1D & 1A + An additional instruction:
        \newline 4. The cleaned code must have the same style and usage of operators as the original provided code. Operators usually start with ``\textbackslash'' such as ``\textbackslash\textless in\textgreater'', ``\textbackslash\textless cdot\textgreater''. Here are some additional Isabelle/ZF code examples which have the same style as the original provided code:
        \newline\{retrieved 3-shot formal statements\}\\
        \bottomrule
    \end{tabular}}
    \caption{Prompts for informalization.}
    \label{tab:prompt_den}
\end{table*}

\begin{table*}[t]
    \centering
    \scalebox{0.88}{
    \begin{tabular}{p{1.1\textwidth}}
        \toprule
        You are an expert in Isabelle theorem prover. You will be provided with an Isabelle/ZF code generated by a language model. The provided code has some Isabelle/ZF syntax errors according to the Isabelle prover. You will also be provided with the error details and where the error code is located in the code. Your task is to fix related errors in the provided Isabelle/ZF code with following instructions. Instructions:\newline
        1. Only refine the code part which is related to provided error details. You must keep other code parts unchanged.\newline
        2. The syntax errors might cause by the mismatch of brackets, incorrect using of operators or invalid representation of Isabelle/ZF code. You should only refine the error codes based on the error details by rewriting, fixing or removing error codes.\newline
        3. You should only output tokens that compose the cleaned code. Anything else, including but not limited to note, description, explanation and comment, must be removed from the final answer. Giving any additional text is prohibited.\newline
        4. The cleaned code must have the same style and usage of operators as the original provided code. Operators usually start with ``\textbackslash'' such as ``\textbackslash\textless in\textgreater'', ``\textbackslash\textless cdot\textgreater''. Here are some additional Isabelle/ZF code examples which have the same style as the original provided code:\newline
        \{retrieved 3-shot formal statements\}\newline
        Strictly follow the instructions that I have claimed.\newline
        Provided Isabelle/ZF Code:\newline
        \{isabelle code\}\newline\{first syntax error details\}\newline
        Refined Code: \\
        \bottomrule
    \end{tabular}}
    \caption{Auto-SEF prompt.}
    \label{tab:prompt_sef}
\end{table*}

\section{Detailed Results}
\label{app:scores}

We provide the exact number of scores of denoising in Table~\ref{tab:round1_emnlp} and Auto-SEF in Table~\ref{tab:round2_lan_1}.
\begin{table*}[t]
    \centering
    \begin{tabular}{l p{0.25\textwidth} c c c c c}
        \toprule
        LLM & Method & BLEU-2 & ChrF & RUBY & CBS & Pass\\
        \hline
        Mistral & Retrieval 3-shot & 6.33 & 48.45 & 28.99 & 76.40 & 17.15\\
        \hline
        Mistral & Retrieval 3-shot+CBD & 10.05 & \textbf{51.38} & 44.82 & \textbf{76.93} & 21.53\\
        \hline
        Mistral & PBD 1A & 8.88 & 38.27 & 38.23 & 68.04 & 6.57\\
        \hline
        Mistral & PBD 1A+CBD & 10.49 & 37.92 & 40.35 & 68.01 & 6.57\\
        \hline
        Mistral & PBD 1B & 11.30 & 43.25 & 42.08 & 70.51 & 9.12\\
        \hline
        Mistral & PBD 1B+CBD & 13.29 & 43.19 & 43.99 & 70.44 & 9.12\\
        \hline
        Mistral & PBD 1C & 15.21 & 44.52 & 44.59 & 71.92 & 13.50\\
        \hline
        Mistral & PBD 1C+CBD & 16.70 & 44.29 & 45.38 & 71.93 & 13.87\\
        \hline
        Mistral & PBD 1D & 14.90 & 48.51 & 46.43 & 74.07 & 28.10\\
        \hline
        Mistral & PBD 1D+CBD & \textbf{17.32} & 48.62 & \textbf{47.41} & 74.10 & 28.10\\
        \hline
        \hline
        GPT-3.5-Turbo & Retrieval 3-shot & 36.06 & \textbf{59.70} & \textbf{58.56} & \textbf{79.34} & \textbf{64.96}\\
        \hline
        GPT-3.5-Turbo & Retrieval 3-shot+CBD& 36.32 & 59.63 & 58.51 & 79.14 & 64.60\\
        \hline
        GPT-3.5-Turbo & PBD 1A & \textbf{38.60} & 57.90 & 58.16 & 78.79 & 63.87\\
        \hline
        GPT-3.5-Turbo & PBD 1A+CBD & 38.59 & 57.86 & 58.12 & 78.63 & 63.87\\
        \hline
        GPT-3.5-Turbo & PBD 1B & 36.49 & 57.08 & 57.79 & 78.27 & 62.04\\
        \hline
        GPT-3.5-Turbo & PBD 1B+CBD & 36.49 & 57.08 & 57.79 & 78.27 & 62.04\\
        \hline
        GPT-3.5-Turbo & PBD 1C & 37.10 & 57.28 & 57.83 & 78.62 & 63.50\\
        \hline
        GPT-3.5-Turbo & PBD 1C+CBD & 37.10 & 57.28 & 57.83 & 78.62 & 63.50\\
        \hline
        GPT-3.5-Turbo & PBD 1D & 38.50 & 58.09 & 58.17 & 78.99 & 64.60\\
        \hline
        GPT-3.5-Turbo & PBD 1D+CBD & 38.50 & 58.09 & 58.17 & 78.99 & 64.60\\
        \bottomrule
    \end{tabular}
    \caption{The effect of denoising.}
    \label{tab:round1_emnlp}
\end{table*}

\begin{table*}[t]
    \centering
    \begin{tabular}{p{0.2\textwidth} p{0.2\textwidth} c c c c c}
        \toprule
        LLM & Method & BLEU-2 & ChrF & RUBY & CBS & Pass\\
        \hline
        Mistral & Iteration1 & 14.91 & 45.69 & 44.16 & 72.22 & 30.66\\
        Mistral & Iteration2 & 13.23 & 44.84 & 43.72 & 72.04 & 32.12\\
        Mistral & Iteration3 & 12.69 & 44.10 & 42.19 & 71.63 & 32.48\\
        Mistral & Iteration4 & 11.29 & 44.18 & 42.30 & 71.53 & 32.85\\
        Mistral & Iteration5 & 11.91 & 43.57 & 41.72 & 71.06 & 33.58\\
        Mistral & Iteration6 & 11.87 & 43.48 & 41.69 & 71.09 & 33.58\\
        Mistral & Iteration7 & 11.72 & 43.64 & 41.26 & 70.91 & 33.58\\
        Mistral & Iteration8 & 11.10 & 43.24 & 41.55 & 71.00 & 33.58\\
        Mistral & Iteration9 & 11.17 & 43.09 & 40.85 & 70.80 & 33.58\\
        \hline
        \hline
        GPT-3.5-Turbo & Iteration1 & 38.11 & 57.66 & 57.45 & 78.71 & 68.98\\
        GPT-3.5-Turbo & Iteration2 & 38.10 & 57.55 & 57.55 & 78.47 & 69.71\\
        GPT-3.5-Turbo & Iteration3 & 38.09 & 57.55 & 57.57 & 78.48 & 70.07\\
        GPT-3.5-Turbo & Iteration4 & 37.99 & 57.54 & 57.50 & 78.45 & 70.44\\
        GPT-3.5-Turbo & Iteration5 & 38.08 & 57.58 & 57.62 & 78.51 & 70.44\\
        GPT-3.5-Turbo & Iteration6 & 38.05 & 57.57 & 57.46 & 78.47 & 70.44\\
        GPT-3.5-Turbo & Iteration7 & 38.00 & 57.53 & 57.39 & 78.49 & 70.44\\
        GPT-3.5-Turbo & Iteration8 & 37.99 & 57.57 & 57.37 & 78.50 & 70.80\\
        GPT-3.5-Turbo & Iteration9 & 38.01 & 57.55 & 57.42 & 78.48 & 70.80\\
        \bottomrule
    \end{tabular}
    \caption{Auto-SEF results with applied CBD.}
    \label{tab:round2_lan_1}
\end{table*}
\end{document}